\documentclass[runningheads]{llncs}

\usepackage{eccv}

\usepackage{eccvabbrv}

\usepackage{amsmath,amssymb} % define this before the line numbering.
\usepackage{diagbox}
\usepackage{booktabs}
\usepackage{makecell}
\usepackage{array,multirow,graphicx}
\usepackage[linesnumbered,ruled,vlined]{algorithm2e}
\usepackage{wrapfig}
\usepackage{xcolor}

\newcommand{\cut}[1]{}

\usepackage{colortbl}
\definecolor{Gray}{gray}{0.9}
\usepackage[accsupp]{axessibility}  % Improves PDF readability for those with disabilities.

\usepackage{hyperref}

\usepackage{orcidlink}

\begin{document}

% ---------------------------------------------------------------
% TODO REVIEW: Replace with your title
\title{Rethinking Unsupervised Outlier Detection via Multiple Thresholding}

% TODO REVIEW: If the paper title is too long for the running head, you can set
% an abbreviated paper title here. If not, comment out.
\titlerunning{Rethinking Unsupervised Outlier Detection via Multiple Thresholding}

% TODO FINAL: Replace with your author list. 
% Include the authors' OCRID for the camera-ready version, if at all possible.
\author{Zhonghang Liu\inst{1}~\orcidlink{0009-0007-6589-7196} \and
Panzhong Lu\inst{2}~\orcidlink{0009-0007-8970-8134} \and
Guoyang Xie\inst{3,4}~\orcidlink{0000-0001-8433-8153} \and Zhichao Lu\inst{4}~\orcidlink{0000-0002-4618-3573} \and Wen-Yan Lin\inst{1}~\orcidlink{0000-0002-1681-6595}}

% TODO FINAL: Replace with an abbreviated list of authors.
\authorrunning{Z. Liu et al.}
% First names are abbreviated in the running head.
% If there are more than two authors, 'et al.' is used.

% TODO FINAL: Replace with your institution list.
\institute{Singapore Management University  \and
Westlake University
 \and  Contemporary Amperex Technology Co., Limited
 \and
City University of Hong Kong \\ 
\email{zhliu.2020@phdcs.smu.edu.sg} \\
\email{lupanzhong@westlake.edu.cn} \\
\email{guoyang.xie@ieee.org} \\
\email{luzhichaocn@gmail.com} \\
\email{daniellin@smu.edu.sg}
}

\maketitle
 
\begin{abstract}
In the realm of unsupervised image outlier detection, assigning outlier scores holds greater significance than its subsequent task: thresholding for predicting labels. This is because determining the optimal threshold on non-separable outlier score
functions is an ill-posed problem. However, the lack of predicted labels not only
hinders some real applications of current outlier detectors but also causes these methods not 
to be enhanced by leveraging the dataset’s self-supervision. To advance
existing scoring methods, we propose a multiple thresholding (Multi-T)
module. It generates two thresholds that isolate
inliers and outliers from the unlabelled target dataset, whereas outliers are employed to obtain better feature representation while inliers provide an uncontaminated
manifold. Extensive experiments verify that Multi-T can significantly improve proposed outlier scoring methods. Moreover, Multi-T contributes to a naive distance-based method being state-of-the-art. Code is available at: \href{https://github.com/zhliu-uod/Multi-T}{https://github.com/zhliu-uod/Multi-T}.

\end{abstract}

% ---------------------------------------------------------------
\section{Introduction}
Which is more important for the unsupervised outlier detection (UOD), \textit{outlier score} or \textit{label}? Currently, the mainstream UOD approaches \cite{lai2019robust, li2022ecod, wang2019effective, 9913887} focus on the first, i.e., learning a discriminative outlier score function. 
However, those recent outlier detectors are usually complex and non-iterative since the absence of predicted labels limits these methods to be further advanced with the dataset's self-supervision. Such that, this work tends to label the unlabelled target dataset, thereby leveraging the dataset's prior knowledge to enhance simple outlier scoring methods\setcounter{footnote}{0}\footnote{Outlier scoring methods refers to the classical outlier detectors.}, achieving state-of-the-art (SOTA) results efficiently.

To understand the importance of the target dataset's self-supervision, we first decompose the predominant image outlier scoring approaches \cite{scholkopf2001estimating, ruff2018deep, lin2021shell, bergman2020classification} into two continuous stages: inlier/normal manifold learning and distance/similarity inference. There is no doubt that identifying inliers will contribute to learning an uncontaminated inlier manifold (e.g. the hypersphere of DeepSVDD \cite{ruff2018deep}). Additionally, motivated by Shell Theory \cite{lin2021shell, lin2023distance} in the high-dimensional space, image feature representation for distance computation can be significantly improved via outliers-based Shell Normalization \cite{lin2021shell}. Such that it is critical to obtain the thresholds that separate both inliers and outliers from target datasets.

\begin{figure*}[t!]
\centering
\includegraphics[width=1.0\textwidth]{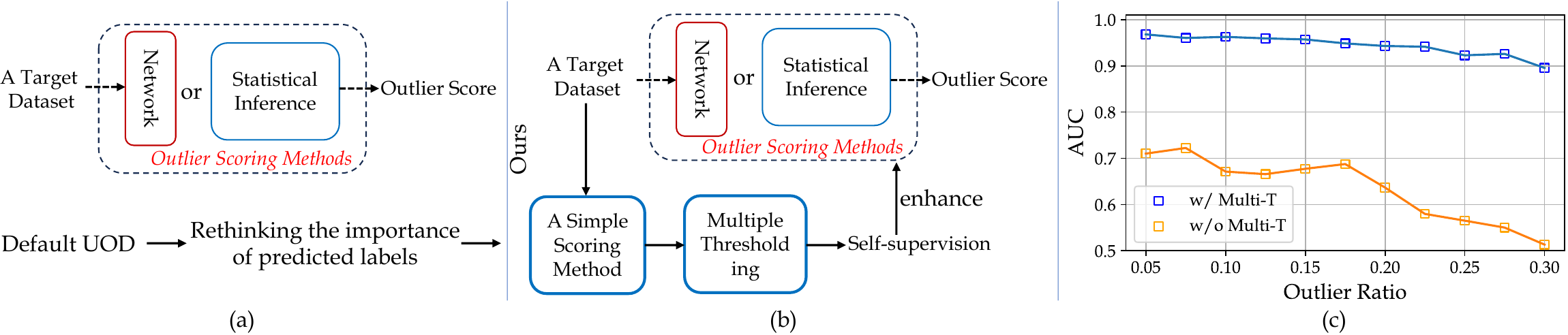} 
\caption{Unlike the default UOD paradigm (a) concerned about learning a discriminative score function, our perspective (b) is to explore the dataset's prior knowledge via thresholding, to advance the previously proposed method. (c) illustrates the significant improvement of DeepSVDD \cite{ruff2018deep} with Multi-T module.
}
\label{figure 1}
\end{figure*}

To align with our above objective, we present a \underline{Multi-T} module, which follows a \textit{multiple} thresholding mechanism.
% , whereas the inlier threshold delineates inliers while the outlier threshold segregates outliers. 
Compared with conventional \textit{single} thresholding methods \cite{ren2018robust, hashemi2019filtering, iouchtchenko2019deterministic, rengasamy2021towards}, Multi-T generates two distinct thresholds that separably isolate outliers and inliers from the target dataset, which is capable of dealing with the inevitable inseparability of the initial given score distribution. This allows us to directly implement Multi-T on a simple outlier score function. 
 
By understanding the impact of the outlier ratio, Multi-T involves two stages: 

\noindent \textbf{{Uncontaminated inliers.}}
We employ an iterative two-step process (\romannumeral1) identifying a noisy inlier distribution through the analysis of sorted initial score function, i.e., Ergodic-set normalized \cite{lin2022locally} distance to the mean of target dataset; (\romannumeral2) subsequently filtering outliers with 3-sigma rule \cite{pukelsheim1994three}. When converged, we can identify uncontaminated inliers, thereby training an inlier/normal manifold.

\noindent {\textbf{Adaptive outliers.}} 
The uncontaminated inliers identification process also outputs a series of outlier threshold candidates. Subsequently, we compare the ranking-index similarity between Ergodic-set Normalization \cite{lin2022locally} and Shell Normalization \cite{lin2021shell} on the initial outlier score function. The structural consistency and contrastive properties of the two normalization procedures motivate us to implicitly estimate a rough outlier ratio, thereby detecting the adaptive outliers.

The Multi-T module is training-free, highly efficient and grants the adaptiveness to choose thresholds suited to downstream implementations. 
As evaluated with comprehensive experimental settings, Multi-T helps to boost significant performance (efficacy and efficiency) improvement. For example, the AUC score of DeepSVDD \cite{ruff2018deep} is improved from $0.622$ to ${0.925}$ on STL-10 dataset \cite{coates2011analysis}. Moreover, the naive distance-to-the-target dataset's mean \cite{lin2021shell} metric integrated with Multi-T achieves SOTA results with only ${1.2}$ second consumption for $10,000$ ResNet-50 \cite{he2016deep} samples that are orders of magnitude faster than baselines.
% \noindent {\textbf{Contributions.}} 
\noindent Our primary contributions are summarized below: \\
\noindent $-$ We introduce a novel perspective for UOD about enhancing previously proposed scoring methods via thresholding on a simple initial outlier score function. \\
\noindent $-$ We present a multiple thresholding (Multi-T) module that generates two different thresholds to separate both inliers and outliers from the target dataset. \\
% It well separate the contaminated target dataset with high efficiency and robustness. \\
\noindent $-$ 
The efficacy of integrated scoring methods can be significantly improved without the external complexity increase, as evaluated in extensive experiments.

% ---------------------------------------------------------------
\section{Related Work}
\label{sec: related works}
\subsection{Unsupervised Outlier Detection}
% \paragraph{\textbf{Inlier-Manifold Learning.}}
Classic UOD task aims to assign an outlier score/likelihood to an
image sample. The recent models can be divided into inlier-manifold learning and outlier exposure.  
Inlier-manifold learning assumes that inliers are the majority. Deep learning-based outlier detectors, as illustrated in studies \cite{lai2019robust,masci2011stacked,chalapathy2017robust,zong2018deep,xia2015learning,zhou2017anomaly,wang2019effective,li2021deep}, typically focus on reducing the dimensionality of image data by projecting it into a latent and discriminative space. Additionally, various statistics-based methods model the inlier manifold using discrimination-based \cite{lin2021shell,lin2022locally} or density-based approaches \cite{breunig2000lof,parzen1962estimation,li2022ecod,kriegel2008angle}.
In applications where outliers are not the minority \cite{wang2019robustness,camposeco2017toroidal}, different approaches are required. Some outlier exposure methods \cite{hendrycks2018deep,liznerski2022exposing} employ out-of-distribution (OOD) data to train networks for detecting unseen OOD samples. For example, Shell-Renormalization \cite{lin2021shell} employs predicted outlier candidates to iteratively refine feature representation. It is particularly useful in scenarios where classic manifold learners, assuming outliers as a minority, may not be effective.
However, these two paradigms are separable, this work will attempt to design a thresholding framework that generates both inliers and outliers, leveraging the benefits of both.
inlier-manifold learning and outlier exposure.

\subsection{Thresholding}
Some traditional outlier detection methods \cite{scholkopf2002learning,liu2008isolation,lin2021shell}, provide both the outlier score and its corresponding threshold concurrently. In this context, the threshold manifests as a hypersphere that accepts the inliers while rejecting those outliers. 
Unquestionably, discerning 
it presents its challenges in high-dimensional space \cite{lin2022locally}. In response, we venture into an alternate perspective. 
% \paragraph{\textbf{Thresholding in One-D Space.}} 
Some other threshold detectors are mostly based on statistical analysis that can be utilized for any given 1-$d$ outlier score function \cite{lai2019robust,lin2021shell,li2022ecod}. 
The representative works involve Kernel-based
\cite{ren2018robust, qi2021iterative}; Curve-based \cite{friendly2013elliptical}; Normality-based: \cite{amagata2021fast, bol1975chauvenet}; Filtering-based \cite{hashemi2019filtering, thanammal2014effective}; Statistical-based \cite{rengasamy2021towards, martin2006evaluation, iouchtchenko2019deterministic, barbado2022rule, bardet2017new, afsari2011riemannian, coin2008testing, alrawashdeh2021adjusted, aggarwal2017introduction, archana2015periodicity, breunig2000lof, van2020evaluation} and Transformation-based \cite{keyzer1997using, raymaekers2021transforming}.
In practice, the default thresholding follows a $\textit{single}$ thresholding paradigm is not realistic since the perfect discriminativeness of the outlier score function is usually not assumed. Therefore, this work introduces a new perspective: $\textit{multiple}$ thresholding.

% ---------------------------------------------------------------
 
\section{Overview}
\subsection{Problem Definition}\label{sec:problem_definition}
Before introducing our method, we first formally define the problem. The target dataset $\mathbb{Z} = \mathbb{Z}_{in} \cup \mathbb{Z}_{out}$ involves $n$ unlabelled images, here $\mathbb{Z}_{in}$ and $\mathbb{Z}_{out}$ represent inliers (normal data) and outliers (anomalous data), respectively. The outlier (contamination) ratio $\gamma  \in (0, 1)$ is denoted as $\frac{\# \mathbb{Z}_{out}}{\# \mathbb{Z}_{in}+\# \mathbb{Z}_{out}}$. Notably, this setup differs from the related semi-supervised outlier detection task by eliminating the need for a prior train/test split for the target dataset. Aligning with previous works \cite{lin2021shell,lin2022locally,lai2019robust}, the raw images are extracted to feature representation $\mathbf{X} = \mathbf{X}_{\text{in}} \cup \mathbf{X}_{\text{out}} = \{\mathbf{x}_{i}\}_{i=1}^{n}$ with pixel representation \cite{lin2022locally, lin2021shell} or pre-trained neural networks (e.g., ResNet \cite{he2016deep} and CLIP \cite{radford2021learning}). 
The main objective of UOD is to create an outlier score function $\text{F}(\cdot)$ to evaluate the likelihood of an image feature $\mathbf{x}_i \in \mathbf{X}$ being outliers:
\begin{equation}
\text{Y}(\mathbf{x}_{i}) = \left\{\begin{array}{ll} 
	0, & \text{F}(\mathbf{x}_{i}) < \phi  \\
	1, & \text{F}(\mathbf{x}_{i}) \geq \phi, 
\end{array}\right.
\label{equ 1} 
\end{equation}
where $\text{Y} = 0 \  (inlier )/1 \ (outlier)$ refers to predicted labels and $\phi$ is the threshold (decision boundary). In this study, we not only center on evaluating the ranking accuracy of $\text{F}(\cdot)$ \cite{lin2021shell, lin2022locally, lai2019robust, ruff2018deep}, but also measure the efficacy of the threshold $\phi$.

\subsection{Rethinking UOD}
As there is no prior train/test split for the target dataset, a classic outlier scoring method $\text{M}(\cdot)$ predicts the outlier score function $\text{F}(\mathbf{X}) = \{ \text{F}(\mathbf{x}_i) \}_{i=1}^{n}$ follows the below process:
\begin{equation}
    \begin{aligned}
        \text{F}(\mathbf{X}) = \text{M}.\operatorname{fit}(\mathbf{X}).\operatorname{predict}(\mathbf{X}),
    \end{aligned}
\label{equ 2}
\end{equation}
where $\operatorname{fit}(\cdot)$ and $\operatorname{predict}(\cdot)$ refer to fitting the target dataset and predicting outlier scores, respectively. Motivated by previous research \cite{lin2021shell, lin2023distance}, Shell normalization $\text{S-norm}(\cdot)$ illustrated below is an \textit{ideal} feature de-noising/refining paradigm. 
\begin{equation}
\begin{aligned}
\text{S-norm}(\mathbf{x}_i, \mathbf{v}_{\text{S}}) = \frac{\mathbf{x}_i-\mathbf{v}_{\text{S}}}{\displaystyle ||\mathbf{x}_i-\mathbf{v}_{\text{S}}||_2}, \mathbf{v}_{\text{S}} = \bigg [\frac{1}{n}\sum_{i=1}^n  \mathbf{X}_{\text{out}} [i][1], \cdots, \frac{1}{n}\sum_{i=1}^n  \mathbf{X}_{\text{out}} [i][d]\bigg],
\label{equ 3}
\end{aligned}
\end{equation}
where $\displaystyle || \cdot ||_2$ refers to $\ell_2\text{-norm}$ and $d$ is the feature dimension. 
% (e.g. $d=2048$ if we use the ResNet feature)
Thus, our first stage is to identify outlier candidates $\mathbf{X}^{\prime}_{\text{out}}$.
% $\mathbf{X}^{\prime}_{\text{out}} = \{x_{i}|\text{F}_{\text{init}}(x_{i}) > \phi_{out} \}$. 
Subsequently, as how some one-class learning-based methods (e.g., OCSVM \cite{scholkopf2001estimating}, DeepSVDD \cite{ruff2018deep}) usually perform, we secondly predict the inliers $\mathbf{X}^{\prime}_{\text{in}}$
% $\mathbf{X}^{\prime}_{\text{in}} = \{x_{i}|\text{F}_{\text{init}}(x_{i}) \leq \phi_{\text{in}} \}$
to fit an inlier/normal manifold. Thus, our objective is to identify both $\mathbf{X}^{\prime}_{\text{in}}$ and $\mathbf{X}^{\prime}_{\text{out}}$.
  
\begin{figure*}[htb]
\centering
\includegraphics[width=1.0\textwidth]{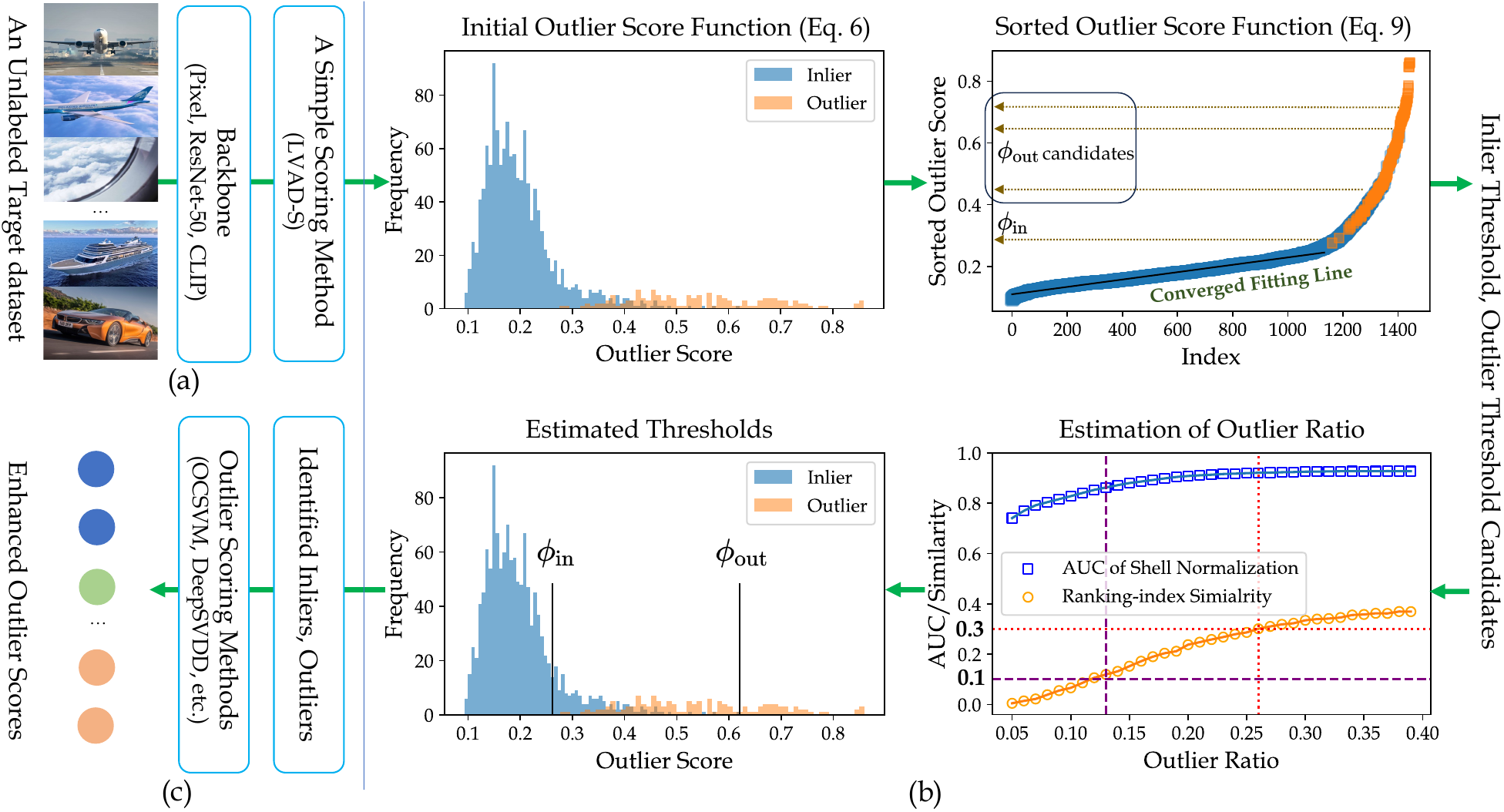}
\caption{The overall paradigm of adopting the multi-thresholds learning (Multi-T) module to advance the existing outlier scoring methods. (a) The preparation consists of feature extraction and an initial outlier score function. (b) Visualization of our Multi-T module. (c) Integrating the predicted inliers and outliers with the previously proposed outlier detectors and obtaining an enhanced outlier score function.
}
\label{figure 2}
\end{figure*}

\begin{algorithm}[t!]
\caption{MTL}
\noindent \KwIn{Initial Outlier Score Function: $\text{F}_{\text{init}}(\mathbf{X})$, Outlier Scoring Method: $\text{M}(\cdot)$}
\noindent \KwOut{Advanced Outlier Score Function: $\text{F}_{\text{M}+\text{Multi-T}}(\mathbf{X})$}
\textcolor{orange}{\tcp{Stage 1: Identifying Uncontaminated Inliers}}
\For{$b$ in iterations}{
    \If{not converged}{
        $\widehat{\text{F}}_{\textit{init}}(\mathbf{X})= \operatorname{sort}(\text{F}_{\text{init}}(\mathbf{X}))$ \ \ \ \ \ \ \ \ \ \ \ \ \ \ \ \ \ \ \ \ \ \   
        \textcolor{red}{\tcp{sort-transform(Eq.\ref{equ 9})}}
        $I^{b} = \left \{\widehat{\text{F}}_{\text{init}}(\mathbf{x}_i) | i< \max_{i} \left \{g(a_i) > \widehat{\text{F}}_{\text{init}}(\mathbf{x}_i) \right \} \right \}$ \
        \textcolor{red}{\tcp{linear-fit(Eq.\ref{equ 10})}}
        $\phi^{b}_{\text{out}} = \operatorname{mean}(I^{b}) + 3\cdot\operatorname{std}(I^{b})$ \ \ \ \ \ \ \ \ \ \
        \textcolor{red}{\tcp{outlier threshold (Eq.\ref{equ 11})}}

        $I^{b+1} = I^{b}.\operatorname{remove}( \{\text{F}_{\text{init}}(\mathbf{x}_i)|\text{F}_{\text{init}}(\mathbf{x}_i) > \phi^{b}_{\text{out}}  \})$ \
        \textcolor{red}{\tcp{filtering(Eq.\ref{equ 11})}}
    }
    $b = b+1$
}
\textbf{Converged} $b: b^*$ \\
$\phi_{\text{in}} = \max (I^{b^{*}})$ \ \ \ \ \ \ \  \ \ \ \ \ \ \ \ \  \ \ \ \ \ \ \ \ \ \ \ \ \  \ \ \ \ \ \ \ \ \ \ \ \
\textcolor{red}{\tcp{inlier threshold(Eq.\ref{equ 12})}}
$\mathbf{X}^{\prime}_{\text{in}} = \{\mathbf{x}_{i}|\text{F}_{\text{init}}(\mathbf{x}_{i}) \leq \phi_{\text{in}} \}$ \ \ \ \ \  \ \ \ \ \ \ \ \ \ \ \ \ \  \ \ \ \ \ \ \  \textcolor{red}{\tcp{predicted inliers(Eq.\ref{equ 13})}}

\textcolor{orange}{\tcp{Stage 2: Identifying Adaptive Outliers}}
$\rho \leftarrow \operatorname{similarity}(\text{F}_{\text{S}}(\mathbf{X}), \text{F}_{\text{E}}(\mathbf{X}))$     \ \ \ \ \
\textcolor{red}{\tcp{outlier ratio estimation(Eq.\ref{equ 15},\ref{equ 16})}}
% \If{$\rho > 0.3$} 
%     {\State $\phi_{\text{out}} = 
%  \phi^{*}_{\text{out}}$} 
% \Else{
%     \If{$ 0.1 \leq \rho \leq 0.3$}{
%             \State $\phi_{\text{out}} = \phi^{1}_{\text{out}}$
%         } 
%     \EndIf
%     \Else{$\phi_{\text{out}} = \phi^{0}_{\text{out}}$}
%     }
% \EndIf 
\eIf{$\rho > 0.3$}{
    $\phi_{\text{out}} = \phi^{*}_{\text{out}}$
}{
    \eIf{$ 0.1 \leq \rho \leq 0.3$}{
        $\phi_{\text{out}} = \phi^{1}_{\text{out}}$
    }{
        $\phi_{\text{out}} = \phi^{0}_{\text{out}}$
    }
}
$\mathbf{X}^{\prime}_{\text{out}} = \{\mathbf{x}_{i}|\text{F}_{\text{init}}(\mathbf{x}_{i}) > \phi_{\text{out}} \}$ \ \  \ \ \ \ \ \ \ \ \ \ \ \ \ \ \ \ \ \ \textcolor{red}{\tcp{predicted outliers(Eq.\ref{equ 20})}}
\textcolor{orange}{\tcp{Stage 3: Adopting Multi-T to Outlier Scoring Method}}
% $\mathbf{v}_{\text{S}}^{\prime} = \operatorname{mean}(\mathbf{X}^{\prime}_{\text{out}})$ \\
% $\text{F}_{\text{M}+\text{Multi-T}}(\mathbf{X}) = \text{M}.fit(\text{S-norm}(\mathbf{X}^{\prime}_{in}, \mathbf{v}_{\text{S}}^{\prime})).predict(\text{S-norm}(\mathbf{X},  \mathbf{v}_{\text{S}}^{\prime})) $ \\

\Return $\text{M}.\operatorname{fit}(\{\text{S-norm}(\mathbf{x}_i, \mathbf{v}_{\text{S}}^{\prime})|\mathbf{x}_i \in \mathbf{X}^{\prime}_{in}\}) .\operatorname{predict}(\{\text{S-norm}(\mathbf{x}_i,  \mathbf{v}_{\text{S}}^{\prime})|\mathbf{x}_i \in \mathbf{X}\})$
\end{algorithm}

\section{Method}
Our solution is displayed as below subsections. Sec.~\ref{sec: 4.1}: we introduce the initial outlier score function; Sec.~\ref{sec: 4.2}: we present the Multi-T module, to generate two thresholds for separating inliers and outliers from the target dataset; 
Sec.~\ref{sec: 4.3}: 
we leverage predicted inliers and outliers to advance previously proposed methods. 

\subsection{Initial Outlier Score Function}
\label{sec: 4.1}
% Before presenting our 
% algorithm in detail, we first formally define the overall process of adopting thresholding for UOD.
% \noindent \textbf{Initial Outlier Score Function.}
In our approach, we employ LVAD \cite{lin2022locally}, one of the UOD SOTAs, as the initial outlier score function. LVAD assumes each image feature $\mathbf{x}_i \in \mathbf{X}$ comes from one of $T$ high-dimensional generative processes.
% \begin{equation}
% \label{eq:gen_para}
% \{a_j, \boldsymbol{\mu}_j   \,|\, j \in \{1, 2, \hdots, m\}\},
% \end{equation}
Thus, the outlier score becomes a sum of the weighted distance of the given instance, $\mathbf{x}_i$,
arising from each generative process $\{w_t, \mathbf{m}_t   \,|\, t \in \{1, \hdots, T\}\}$, which can be simplified as:
\begin{equation} 
\text{F}_{\text{LVAD}}(\mathbf{x}_i)  =  \sum_{t=1}^T w_t \cdot \text{\text{Dist}}(\text{E-norm}(\mathbf{x}_i), \text{E-norm}(\mathbf{m}_t)),
\label{equ 4}
\end{equation}
where $\text{\text{Dist}}(\cdot)$ refers to the $\ell_2\text{-norm}$ metric (Euclidean distance), $w_t$ is the weight of the $t$-th generative process,  $\mathbf{m}_t$ is the $t$-th generative process's centroid and $\text{E-norm}(\cdot)$ refers to the Ergodic-set normalization, an effective and outlier ratio invariant normalization procedure, illustrated as follows:
\begin{equation}
\begin{aligned}
\text{E-norm}(\mathbf{x}_i, \mathbf{v}_{\text{E}}) =  \frac{\mathbf{x}_i-\mathbf{v}_{\text{E}}}{\displaystyle ||\mathbf{x}_i-\mathbf{v}_{\text{E}}||_2} ,
 \mathbf{v}_{\text{E}} = \bigg[ \frac{1}{n \cdot d}\sum_{i=1}^n \sum_{j=1}^d \mathbf{x}_{i,j}, \cdots,  \frac{1}{n \cdot d}\sum_{i=1}^n \sum_{j=1}^d \mathbf{x}_{i,j} \bigg],
\end{aligned} 
\label{equ 5}
\end{equation}
where $\mathbf{v}_{\text{E}}$ is its reference vector and $j$ is the dimension index.
To gain higher efficiency and maintain efficacy, we set $T=1$, such that the initial outlier score function (LVAD-S) is formulated as:
\begin{equation}
\begin{aligned}
\text{F}_{\text{init}}(\mathbf{X})   = \text{F}_{\text{LVAD-S}}(\mathbf{X}) & = \{\text{\text{Dist}}( \text{E-norm}(\mathbf{x}_i, \mathbf{v}_{\text{E}}), \text{E-norm}(\mathbf{m}_{\mathbf{X}}, \mathbf{v}_{\text{E}}))\}_{i=1}^{n}, \\
\mathbf{m}_{\mathbf{X}} & = \bigg[\frac{1}{n}\sum_{i=1}^n \mathbf{x}_{i,1},\cdots,\frac{1}{n}\sum_{i=1}^n \mathbf{x}_{i,d}\bigg],
\label{equ 6}
\end{aligned}
\end{equation}
where $\mathbf{m}_{\mathbf{X}}$ is the mean of target dataset's features.
Despite its simplicity, Eq. \ref{equ 6} is still a reliable outlier scoring method \cite{lin2022locally}.
 
\subsection{Multi-T: Multiple Thresholding}
\label{sec: 4.2}
% \noindent \textbf{Overview.}

% The objective of thresholding in our perspective is to split the target dataset into two components:
% $\mathbf{X}^{\prime}_{\text{in}}$ and $\mathbf{X}^{\prime}_{\text{out}}$. 
\noindent {\textbf{Motivation.}}
In the field of UOD, the conventional thresholding paradigm is to learn a \textit{single} threshold $\phi$ on the initial outlier score function $\text{F}_{\text{init}}(\mathbf{X})$.
% \begin{equation}
% \mathbf{X}^{\prime}_{\text{in}} = \{\mathbf{x}_{i}|\text{F}_{\text{init}}(\mathbf{x}_{i}) \leq \phi \}, \mathbf{X}^{\prime}_{\text{out}} = \{\mathbf{x}_{i}|\text{F}_{\text{init}}(\mathbf{x}_{i}) > \phi \} 
% \label{equ 6}
% \end{equation}
However, it ignores the impact of the outlier ratio $\gamma$. As the mean of target dataset $\mathbf{m}_{\mathbf{X}}$ will be shifted to outliers when $\gamma$ becomes higher, i.e., mean-shift problem \cite{reiss2023mean}, there will be an inevitable overlap between inlier and outlier score distributions. So we define the distribution of the initial outlier score function $D$ as follows:
\begin{equation}
D =   \text{F}_{\text{init}}(\mathbf{X})  =  \text{F}_{\text{init}}(\mathbf{X}_{\text{in}})   \cup   \text{F}_{\text{init}}(\mathbf{X}_{\text{out}})   = I \cup A  \cup O,
\label{equ 7}
\end{equation}
where $I$ and $O$ refer to the uncontaminated inlier and outlier score distributions, $A$ refers to the overlap. 
% Additional, the outlier score distribution $\mathcal{O} = \text{F}_{\text{init}}(\mathbf{x}_{\text{out}})$ is also dependent with $\gamma$.
Such that we are concerned about \textit{multi}-thresholding instead of the
conventional way that explicitly finding a \textit{single} threshold between $\text{F}_{\text{init}}(\mathbf{X}_{\text{in}})$ and $\text{F}_{\text{init}}(\mathbf{X}_{\text{out}})$. Specifically, our objective is identifying the inlier threshold $\phi_{\text{in}}$ and outlier threshold $\phi_{\text{out}}$ that isolate $I$ and $O$ from $D$.
% \begin{equation}
% \mathbf{X}^{\prime}_{\text{in}} = \{\mathbf{x}_{i}|\text{F}_{\text{init}}(\mathbf{x}_{i}) \leq \phi_{\text{in}} \}, \mathbf{X}^{\prime}_{\text{out}} = \{\mathbf{x}_{i}|\text{F}_{\text{init}}(\mathbf{x}_{i}) > \phi_{\text{out}} \} 
% \label{equ 8}
% \end{equation}

\noindent {\textbf{Identifying uncontaminated inliers.}}
We identify $I$ by understanding the initial outlier score function (Eq. \ref{equ 6}), which computes the normalized $\ell_2\text{-norm}$ of each instance $\mathbf{x}_i \in \mathbf{X}$ relative to the mean of $\mathbf{X}$. 
Inspired by Shell Theory \cite{lin2021shell}, $I$ satisfies a Gaussian-like distribution, 
% So before tackling with $S(\mathbf{x})$, we analyze Gaussian distribution, a general assumption of inlier distribution.
the classic statistical estimation of $I$ is:
\begin{equation}
I = \left \{\text{F}_{\text{init}}(\mathbf{x}_i) | \text{F}_{\text{init}}(\mathbf{x}_i) < \operatorname{mean}(D) + k \cdot \operatorname{std}(D) \right \},
\label{equ 8}
\end{equation}
where $\operatorname{mean}(D)$ and $\operatorname{std}(D)$ refer to the mean and standard deviation, $k$ is the hyper-parameter of the "$k$-sigma" rule.
In practice, optimizing $k$ without supervision is challenging. Thus, we convert to a non-parametric way by utilizing an ascending sort 
projection $\operatorname{sort}(\cdot)$ to $\text{F}_{\text{init}}(\mathbf{X})$, which projects $D$ into a 2-$d$ space ($x$-axis: instance index, $y$-axis: sorted outlier score), the sorted outlier score is:
\begin{equation}
\widehat{\text{F}}_{\text{init}}(\mathbf{X}) = \operatorname{sort}(\widehat{\text{F}}_{\text{init}}(\mathbf{X})), i.e., \widehat{\text{F}}_{\text{init}}[i+1]>\widehat{\text{F}}_{\text{init}}[i].
\label{equ 9}
\end{equation}
Intuitively, $\widehat{I} = \operatorname{sort} (I)$ can be fitted with a naive linear regressior $g(a) = \boldsymbol{\beta}^{\top} \cdot a = \beta_0 + \beta_1 \cdot a$, as shown in Fig. \ref{figure 2} (b). So we identify the potential inliers as:
\begin{equation}
\begin{aligned}
    & I = \left \{\widehat{\text{F}}_{\text{init}}(\mathbf{x}_i) | i< \max_{i} \left \{g(a_i) > \widehat{\text{F}}_{\text{init}}(\mathbf{x}_i) \right \} \right \}, s.t. \min _{\boldsymbol{\beta}}\displaystyle ||a^{\top} \boldsymbol{\beta}-\widehat{\text{F}}_{\text{init}}(\mathbf{X})||_2,
\label{equ 10}
\end{aligned}
\end{equation}
where $\boldsymbol{\beta}$ is the coefficients. Inevitably, this fitting process will be shifted with outliers, which is addressed together with the following outliers' identification.

\noindent {\textbf{Identifying adaptive outliers.}}
In statistics, 3-sigma rule \cite{pukelsheim1994three} declares variables $C$ has a large probability
within three standard deviations from the mean 
% $\mu$, i.e. $P(|A-\mu| \leq 3 \sigma) \approx 0.997$, 
if $C$ follows a Gaussian-like distribution. So we employ it with two aspects: \\
\noindent
(\romannumeral1) iteratively filter outliers during the phase of inliers identification:
\begin{equation}
\begin{aligned}
\phi^{b}_{\text{out}} = \operatorname{mean}(I^{b}) + 3\cdot\operatorname{std}(I^{b}),
    I^{b+1} = I^{b}.\operatorname{remove}( \{\text{F}_{\text{init}}(\mathbf{x}_i)|\text{F}_{\text{init}}(\mathbf{x}_i) > \phi^{b}_{\text{out}}  \}),
\end{aligned}
\label{equ 11}
\end{equation}
where $b$ refers to the $b$-th iteration. 
When converged, the inlier threshold is: 
\begin{equation}
   \phi_{\text{in}} = \max (I^{b^*}),
\label{equ 12}
\end{equation}
where $I^{b^*}$ refers to the converged inlier prediction and we can identify inliers as:
\begin{equation}  \mathbf{X}^{\prime}_{\text{in}} = \{\mathbf{x}_{i}|\text{F}_{\text{init}}(\mathbf{x}_{i}) \leq \phi_{\text{in}} \}.
\label{equ 13}
\end{equation}
\noindent (\romannumeral2) During the above phase, we estimate the outlier threshold candidates:
\begin{equation}
    \phi^{1}_{\text{out}} = \operatorname{mean}(I^1) + 3\cdot\operatorname{std}(I^1), \phi^{*}_{\text{out}} = \operatorname{mean}(I^{b^*}) + 3\cdot\operatorname{std}(I^{b^*}),
\label{equ 14}
\end{equation}
where $\phi^{1}_{\text{out}}$, $\phi^{*}_{\text{out}}$ refer to the outlier threshold candidates at the first and final iterations, respectively, i.e., $\phi^{1}_{\text{out}}>\phi^{*}_{\text{out}}$.
% To tackle this problem, 
To obtain a $\gamma$-adaptive outlier threshold, we consider the impact and relationship of two mentioned normalization procedures: Shell Normalization $\text{S-norm}(\cdot)$ (Eq. \ref{equ 3}) and Ergodic-set Normalization $\text{E-norm}(\cdot)$ (Eq. \ref{equ 5}). Both them follow a similar normalization formula (denominator: $\ell_2\text{-norm}$), the difference is their reference vectors $\mathbf{v}^{\prime}_{\text{S}}$ and $\mathbf{v}_{\text{E}}$, contrast as:
\begin{equation}
\begin{aligned}
\mathbf{v}^{\prime}_{\text{S}} & = \bigg [\frac{1}{n}\sum_{i=1}^n \mathbf{X}_{\text{out}}^{\prime}[i][1], \cdots, \frac{1}{n}\sum_{i=1}^n \mathbf{X}_{\text{out}}^{\prime}[i][d]\bigg], \\
\mathbf{v}_{\text{E}} & = \bigg [ \frac{1}{n \cdot d}\sum_{i=1}^n \sum_{j=1}^d \mathbf{x}_{i,j}, \cdots, \frac{1}{n \cdot d}\sum_{i=1}^n \sum_{j=1}^d \mathbf{x}_{i,j} \bigg ].
\label{equ 15}
\end{aligned}
\end{equation}
Obviously, Shell normalization is strongly subjective to the efficacy of outlier prediction $\mathbf{X}^{\prime}_{\text{out}}$ while Ergodic-set Normalization is independent with $\gamma$ that is a sub-optimal but stable operation.
% \noindent \textbf{Estimating Contaminated Ratio $\gamma$.}
To compare these two normalization procedures, we first employ them to Eq. \ref{equ 6}, and obtain two outlier score functions:
% \footnote{For implementation, we integrate Shell normalization with the Robust-Least-Square paradigm, to align with its original paper.}:
\begin{equation}
\begin{aligned}
    \text{F}_{\text{S}}(\mathbf{X}) & = \{\text{Dist}(  \text{S-norm}(\mathbf{x}_i, \mathbf{v}^{\prime}_{\text{S}})  , \text{S-norm}( \mathbf{m}_{\mathbf{X}}, \mathbf{v}^{\prime}_{\text{S}}))\}_{i=1}^{n}, \\
    \text{F}_{\text{E}}(\mathbf{X}) & = \{\text{Dist}( \text{E-norm}(\mathbf{x}_i, \mathbf{v}_{\text{E}}) , \text{E-norm}( \mathbf{m}_{\mathbf{X}}, \mathbf{v}_{\text{E}})) \}_{i=1}^{n},
\end{aligned}
\label{equ 16}
\end{equation}
where $\mathbf{v}^{\prime}_{\text{S}}$ comes from $\mathbf{X}_{\text{out}}^{\prime}$ obtained by MAD \cite{archana2015periodicity} (Shell Normalization's default thresholding method that is only effective on high-$\gamma$). Subsequently, two outlier score functions $\text{F}_{\text{S}}(\mathbf{X})$ and $\text{F}_{\text{E}}(\mathbf{X})$ are arranged in ascending order, and we denote the ranked-index lists as $\mathcal{R}_{\text{S}}$ and $\mathcal{R}_{\text{E}}$. 
% In these lists, the outlier score corresponding to the $i$-th index will be less than that of the $(i+1)$-th index. 
The outlier ratio $\gamma$ can be approximately estimated with the similarity between $\mathcal{R}_{\text{S}}$ and $\mathcal{R}_{\text{E}}$ since $\text{F}_{\text{S}}(\mathbf{X})$ and $\text{F}_{\text{E}}(\mathbf{X})$ share a consistent structure (distance-to-the-mean), making $\mathcal{R}_{\text{S}}$ and $\mathcal{R}_{\text{E}}$ comparable. Besides,  $\text{F}_{\text{E}}(\mathbf{X})$ serves as a reliably effective baseline. When $\gamma$ is low, $\text{S-norm}(\mathbf{X}, \mathbf{v}^{\prime}_{\text{S}})$ tends to under-perform, lead the similarity between $\mathcal{R}_{\text{S}}$ and $\mathcal{R}_{\text{E}}$ becomes low. In contrast, with a high 
$\gamma$, $\text{S-norm}(\mathbf{X}, \mathbf{v}^{\prime}_{\text{S}})$ merely refines the ranking of a small number of outliers, indicative of high similarity.
The similarity $\rho \in [-1, 1]$ is computed with the Pearson correlation coefficient:
\begin{equation}
\rho = \frac{\operatorname{cov}(\mathcal{R}_{\text{S}}, \mathcal{R}_{\text{E}})}{\operatorname{std}(\mathcal{R}_{\text{S}}) \cdot
\operatorname{std}(\mathcal{R}_{\text{E}})},
\label{equ 17}
\end{equation}
where
$\operatorname{cov}(\mathcal{R}_{\text{S}}, \mathcal{R}_{\text{E}})$
and $\operatorname{std}(\mathcal{R}_{\text{S}})/ \operatorname{std}(\mathcal{R}_{\text{E}})$ 
are the covariance and standard deviation of the rank variables.
If the similarity is relatively low ($< 0.1$), we compute the 3-sigma of the entire outlier score distribution $D$ as the outlier threshold:
\begin{equation}
    \phi^{0}_{\text{out}} = \operatorname{mean}(D) + 3\cdot\operatorname{std}(D).
\label{equ 18}
\end{equation}
It is a conservative estimator for outliers since the above linear fitting might suffer from some challenges at low-$\gamma$ since there is no explicit outlier score distribution.
If the similarity is large ($> 0.3$), we choose the relatively smooth outlier threshold $\phi^{*}_{\text{out}}$, while $\phi^{1}_{\text{out}}$ is used for middle-$\gamma$ cases.
The outlier threshold is described as:

\begin{equation}
\phi_{\text{out}} = 
    \left\{\begin{array}{ll}
\phi^{*}_{\text{out}} \ , & \ \operatorname{if} \rho > 0.3   \\
\phi^{1}_{\text{out}} \ , & \ \operatorname{if} 0.1 \leq  \rho \leq 0.3   \\
\phi^{0}_{\text{out}} \ , & \ \operatorname{otherwise}\\
\end{array}\right.
\label{equ 19}
\end{equation} 
Therefore, we can subsequently identify outlier ratio invariant outliers as:
\begin{equation}
\mathbf{X}^{\prime}_{\text{out}} = \{\mathbf{x}_{i}|\text{F}_{\text{init}}(\mathbf{x}_{i}) > \phi_{\text{out}} \}.
\label{equ 20}
\end{equation}
 
\subsection{Implementation}
\label{sec: 4.3}
Firstly, we can directly employ Multi-T with the distance-to-the-mean paradigm:
% of adopting Multi-T to the UOD method $\text{M}(\cdot)$ 
% is defined as follows:
% learning a normal manifold and estimating the outlier likelihood respectively.
\begin{equation}
    \begin{aligned}
\text{F}_{\text{Multi-T}}(\mathbf{X}) & = \{\text{Dist}(\text{S-norm}(\mathbf{x}_i, \mathbf{v}^{\prime}_{\text{S}}), \text{S-norm}( \mathbf{m}_{\mathbf{X}^{\prime}_{\text{in}}}, \mathbf{v}^{\prime}_{\text{S}}))\}_{i=1}^{n}, \\  
\mathbf{v}_{\text{S}}^{\prime} & = \bigg [\frac{1}{n}\sum_{i=1}^n  \mathbf{X}^{\prime}_{\text{out}}[i][1], \cdots, \frac{1}{n}\sum_{i=1}^n  \mathbf{X}^{\prime}_{\text{out}}[i][d]\bigg], \\
\mathbf{m}_{\mathbf{X}^{\prime}_{\text{in}}} & =  \bigg[\frac{1}{n}\sum_{i=1}^n \mathbf{X}^{\prime}_{\text{in}}[i][1],\cdots,\frac{1}{n}\sum_{i=1}^n \mathbf{X}^{\prime}_{\text{in}}[i][d]\bigg].
    \end{aligned}
\label{equ 21}
\end{equation}
Secondly, Multi-T can be integrated with UOD methods $\text{M}(\cdot)$, illustrated as:

\begin{equation}
\begin{aligned}
\text{F}_{\text{M}+\text{Multi-T}}(\mathbf{X})  = \text{M} &.\operatorname{fit}(\{\text{S-norm}(\mathbf{x}_i, \mathbf{v}_{\text{S}}^{\prime})|\mathbf{x}_i \in \mathbf{X}^{\prime}_{in}\}) \\
& .\operatorname{predict}(\{\text{S-norm}(\mathbf{x}_i,  \mathbf{v}_{\text{S}}^{\prime})|\mathbf{x}_i \in \mathbf{X}\}). 
% \\
% \mathbf{v}_{\text{S}}^{\prime} & = \bigg [\frac{1}{n}\sum_{i=1}^n  \mathbf{X}^{\prime}_{\text{out}}[i][1], \cdots, \frac{1}{n}\sum_{i=1}^n  \mathbf{X}^{\prime}_{\text{out}}[i][d]\bigg]
\end{aligned}
\label{equ 22}
\end{equation}

\section{Experiments}
We cover our experiments for (\romannumeral1) comparing with other threshold learners on dataset splitting (Sec.~\ref{sec 5.2}); (\romannumeral2) measuring the performance improvement of adopting the Multi-T to previously proposed outlier scoring methods (Sec.~\ref{sec 5.3}).
% \subsection{Common Settings}

\noindent {\textbf{Benchmarks and Feature Extraction.}}
Based on the existing research \cite{lin2022locally,lin2021shell}, we use the raw pixel representation for grayscale datasets, including MNIST \cite{lecun-mnisthandwrittendigit-2010} and Fashion-MNIST \cite{xiao2017online}. For RGB datasets such as STL-10 \cite{coates2011analysis}, Internet \cite{lin2021shell}, CIFAR-10 \cite{krizhevsky2009learning},   MIT-Places-Small \cite{zhou2017places}, we adopt two deep feature extractors: ImageNet pretrained \cite{he2019rethinking} ResNet-50  \cite{he2016deep} (ResNet)\footnote{Unless otherwise stated, ResNet-50 is the default feature extractor.}, and CLIP \cite{radford2021learning}. 

\noindent {\textbf{Target Dataset.}}
In our experiments, each class within a benchmark dataset is alternatively regarded as inliers, with instances from all other classes considered outliers. The target dataset includes all inliers along with a randomly selected subset of outliers. Results for each dataset are averaged across all classes.
Additionally, for every round, we further average the results over a wide range of outlier ratios: $[0.05, 0.1, 0.2, 0.3, 0.4]$\footnote{Unless otherwise stated, the experimental results are averaged across $[0.05, \cdots, 0.4]$.}, to assess the method's robustness/stability. 
% In the case of CatsVsDogs \cite{elson2007asirra}, which includes only two classes, the outlier ratios are modified to $[0.001, 0.01, 0.1, 0.2, 0.3]$.
\begin{figure*}[htb!]
\centering
\includegraphics[width=1.0\textwidth]{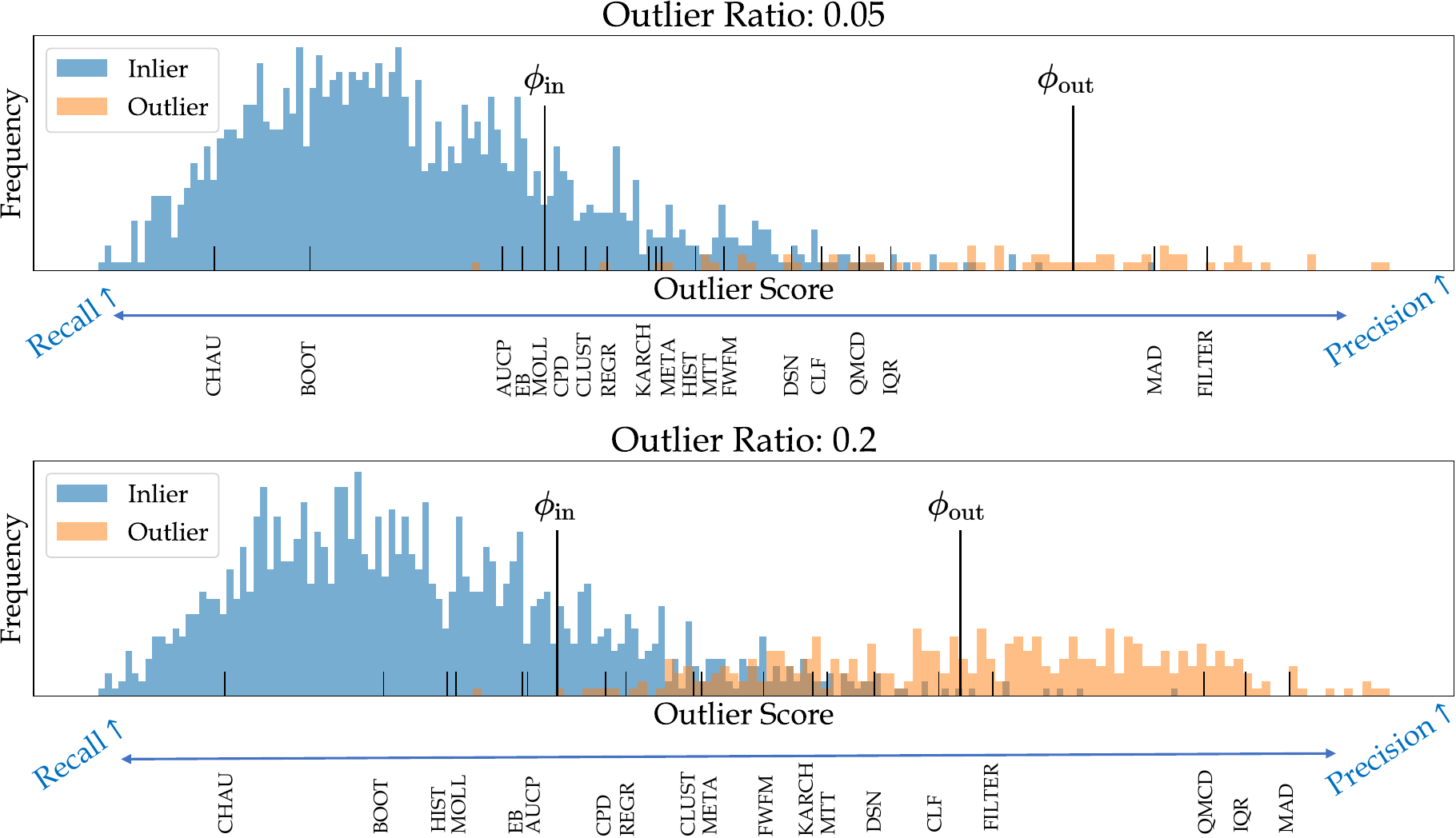} 
\caption{Qualitative results compared with SOTA threshold learners. The experiments are conducted on STL-10 (Inlier class: Monkey). 
}
\label{figure 3}
\end{figure*}

\subsection{Threshold Learning}
\label{sec 5.2}

\noindent 
{\textbf{Competing Methods.}}
We compare our proposed Multi-T with those effective thresholding (TL) methods involving Kernel-based: AUCP \cite{ren2018robust},
FGD \cite{qi2021iterative}; Curve-based: EB \cite{friendly2013elliptical}; Normality-based: DSN \cite{amagata2021fast},  CHAU \cite{bol1975chauvenet}; Filtering-based: FILTER \cite{hashemi2019filtering}, HIST \cite{thanammal2014effective}; Statistical-based: MTT \cite{rengasamy2021towards}, BOOT \cite{martin2006evaluation}, QMCD \cite{iouchtchenko2019deterministic},
CLF \cite{barbado2022rule},
IQR \cite{bardet2017new},
KARCH \cite{afsari2011riemannian},
MCST \cite{coin2008testing},
GESD \cite{alrawashdeh2021adjusted},
REGR \cite{aggarwal2017introduction},
MAD \cite{archana2015periodicity}, CLUST\cite{breunig2000lof},
CPD\cite{van2020evaluation} and Transformation-based: MOLL \cite{keyzer1997using}, YJ \cite{raymaekers2021transforming}.

\noindent
{\textbf{Evaluation Metric.}}
The performance of general thresholding is measured with $\text{F}_\beta$-score, defined as follows:
\begin{equation}
\text{F}_\beta=\left(1+\beta^2\right)\cdot \frac{\text {precision}\cdot \text{recall}}{\left(\beta^2 \cdot \text{precision}\right)+\text{recall}},
\end{equation}
where $\beta<1$ tilts towards precision while $\beta > 1$ prioritizes recall. In this work, we utilize $\text{F}_{0.1}$-score  ($\text{F}_{0.1}$) and $\text{F}_{10}$-score ($\text{F}_{10}$) that measure the accuracy of predicted outliers and inliers. 
Besides, $\text{F}_{1}$-score ($\beta=1$), a harmonic mean of the precision and recall, is not practical for this task since the compete separability between inliers' and outliers' score distributions cannot be guaranteed. 

\begin{table}[htb!]
\caption{Average $\text{F}_{\beta}$-score of thresholding methods on STL-10. "+GT Norm" refers to the ideal initial outlier score function. "$\text{Highest} \  \text{F}_{\beta}$" ($\beta$: 0.1, 10) refers to the method that garners the highest $\text{F}_{\beta}$-score across all baseline models. The better average result is highlighted in \textbf{bold}.
% It's noteworthy that our method consistently rivals state-of-the-art techniques across diverse evaluation metrics. In contrast, many baseline methods exhibit specialization, excelling predominantly in one metric.
}
\footnotesize
\centering
\renewcommand{\arraystretch}{1.2}
\setlength{\tabcolsep}{1.5pt}
{\begin{tabular}{l|l|ccc|ccc}
    \toprule
    % \hline
    \multirow{2}{*}{Outlier Score Fun.}  & \multirow{2}{*}{Method (TL)} & \multicolumn{3}{c}{ResNet-50} & \multicolumn{3}{c}{CLIP} \\
    % \cmidrule(lr){3-5} \cmidrule(lr){6-8}
    \cline{3-8}
    & & \makecell[c]{$\text{F}_{0.1}$} & \makecell[c]{$\text{F}_{10}$} & Avg. & \makecell[c]{$\text{F}_{0.1}$} & \makecell[c]{$\text{F}_{10}$}  & Avg. \\
    \midrule
    % \hline
    \multirow{3}{*}{LVAD-S \cite{lin2022locally}}   &  Highest $\text{F}_{0.1}$ &  0.911  &  0.454 & 0.682  &  0.936  & 0.401  & 0.669    \\
    \multirow{3}{*}{} & Highest $\text{F}_{10}$  &  0.382  & 0.967 & 0.674 & 0.357 & 0.954  & 0.655  \\
    % \cline{2-4}
    \multirow{3}{*}{}  &\cellcolor{gray!10}Multi-T(Ours)     &  \cellcolor{gray!10} 
 0.840 & \cellcolor{gray!10}  0.869 & \cellcolor{gray!10} 
 \textbf{0.855} &  \cellcolor{gray!10} 0.860   & \cellcolor{gray!10} 0.866  & \cellcolor{gray!10} \textbf{0.863} \\
    % \cline{1-6}
    % \multirow{3}{*}{OCSVM \cite{scholkopf2002learning}}   & Highest $\text{F}_{0.1}$   &    &   &    &   \\
    % \multirow{3}{*}{}  & Highest $\text{F}_{10}$   &    &   &    &   \\
    % % \cline{2-4}
    % \multirow{3}{*}{}  &  Multi-T(Ours)   &    &   &    &   \\
    \cline{1-8}
    \multirow{3}{*}{$+$GT Norm \cite{lin2021shell}}   &    Highest $\text{F}_{0.1}$  & 0.976  &  0.452  &  0.714  &  0.977 &  0.448  & 0.713  \\
    \multirow{3}{*}{}  & Highest $\text{F}_{10}$    & 0.447  &  0.979  &  0.713  & 0.691  & 0.986   & 0.839  \\
    % \cline{2-4}
    \multirow{3}{*}{} & \cellcolor{gray!10}Multi-T(Ours)   & \cellcolor{gray!10} 0.917  & 
\cellcolor{gray!10} 0.980   & \cellcolor{gray!10}  \textbf{0.949}  & \cellcolor{gray!10} 0.857  & \cellcolor{gray!10}  0.984  & \cellcolor{gray!10} \textbf{0.920}  \\
    % \cline{1-6}
    % \multirow{3}{*}{OCSVM \cite{scholkopf2002learning}+GT}   & Highest $\text{F}_{0.1}$   &    &   &    &   \\
    % \multirow{3}{*}{}  & Highest $\text{F}_{10}$   &    &   &    &   \\
    % % \cline{2-4}
    % \multirow{3}{*}{}  &  Multi-T(Ours)     &    &   &    &   \\
    \bottomrule
    % \hline
\end{tabular}
 }
\label{table 1}
\end{table}

\noindent {\textbf{Main Results.}}
Across a series of competing threshold learners, while one method may achieve a commendable $\text{F}_{0.1}$-score of $0.976$, the corresponding $\text{F}_{10}$-score is mere $0.452$, and vice versa ($\text{F}_{10}$-score: $0.979$; $\text{F}_{0.1}$-score: $0.447$). By contrast, our solution maintains its efficacy across various evaluation criteria ($\text{F}_{0.1}$-score: $0.917$, $\text{F}_{10}$-score: $0.980$). 
It exhibits remarkable stability across a diverse spectrum of outlier ratios, as illustrated in Fig. \ref{figure 3}.
To further investigate the benefits of our technique, Tab. \ref{table 1} categorizes two groups, each representing a different outlier score function. Such categorization is pivotal as threshold methods ought to accommodate various foundational bases.

{
\renewcommand{\arraystretch}{1.2}
\begin{table}[ht!]
\centering
\footnotesize
\caption{Average AUC results compared with SOTA outlier scoring methods (outlier detectors). \textcolor{blue}{{Blue}} and \textcolor{orange}{{Orange}} indicates the best and second-best results, respectively.
% By integrating with the Multi-T module, the outlier scoring performance of previously proposed methods, e.g., DeepSVDD, Shell-Re and OCSVM has been significantly improved.
} 
\setlength{\tabcolsep}{1.3pt}{
\begin{tabular}
{ll|cc|cc|cc|cc|c}
% {ll|llllllllll}
\toprule
% \hline
& \multirow{2}{*}{{Method (OS)}} &\multicolumn{2}{c}{STL-10} & \multicolumn{2}{c}{CIFAR-10} & \multicolumn{2}{c}{CIFAR-100} & \multicolumn{2}{c}{MIT-Places} & \multicolumn{1}{c}{MNIST}  \\
\cmidrule(lr){3-4} \cmidrule(lr){5-6} \cmidrule(lr){7-8} \cmidrule(lr){9-10} \cmidrule(lr){11-11}  
% \cline{3-11}
& & \makecell[c]{ResNet} & \makecell[c]{CLIP}  & \makecell[c]{ResNet} & \makecell[c]{CLIP} & \makecell[c]{ResNet} & \makecell[c]{CLIP} & \makecell[c]{ResNet} & \makecell[c]{CLIP}  & \makecell[c]{Pixel}  \\
\midrule
% \hline
& IF  \cite{liu2008isolation}     & 0.836 & 0.943  & 0.780 & 0.891  & 0.790 & 0.866 & 0.687 & 0.868  & 0.776 \\
& LOF  \cite{breunig2000lof}      & 0.628 & 0.626  & 0.673 & 0.621  & 0.839 & 0.849 & 0.556 & 0.520  &  0.754 \\
% & VAE     &  &  &   &  &   &  &  &   & \\
% & MOGAAL     &  &  &   &  &   &  &   &   &       \\
% & VAE \cite{kingma2013auto} &  &  &   &  &   &  &   &   &    \\
% & DAGMM \cite{}    &  &  &   &  &   &  &   &   &    \\
% & ADGAN \cite{} &  &  &   &  &   &  &   &   &    \\
& RSRAE  \cite{lai2019robust}   & \textcolor{orange}{{0.962}} & 0.938   & 0.862 & 0.879 & 0.914 & 0.885 & \textcolor{blue}{{0.874}} & 0.876 & 0.793  \\ 
& ECOD  \cite{li2022ecod}   & 0.907 & \textcolor{orange}{{0.981}}  & \textcolor{orange}{{0.873}} &  \textcolor{orange}{{0.935}} & 0.873 & \textcolor{orange}{{0.918}} & 0.777 & \textcolor{orange}{{0.943}}  & 0.734  \\
& LUNAR  \cite{goodge2022lunar}   & 0.776 & 0.821  & 0.767 & 0.774 & 0.838 & 0.899 & 0.643 & 0.871  & 0.797 \\
&  Shell-Re. \cite{lin2021shell} & 0.862 & 0.838 & 0.860 & 0.813 & 0.835 & 0.813 & 0.826 & 0.914  & 0.776 \\
& LVAD  \cite{lin2022locally}  & 0.954 &  0.968 & 0.860 & 0.917 & \textcolor{orange}{{0.921}} & 0.917 & 0.844 & 0.919  & \textcolor{orange}{{0.867}}  \\
% & INNE  &  &  &   &  &   &   &  &  &   &   &      \\
% & VAE  &  &  &   &  &   &   &  &  &   &   &      \\
% & DAGMM  &  &  &   &  &   &   &  &  &   &   &      \\ 
\rowcolor{gray!10} & Multi-T  & \textcolor{blue}{{0.968}} & \textcolor{blue}{{0.989}} & \textcolor{blue}{{0.895}} & \textcolor{blue}{{0.957}} & \textcolor{blue}{{0.938}} & \textcolor{blue}{{0.956}} & \textcolor{orange}{{0.867}} & \textcolor{blue}{{0.974}} & \textcolor{blue}{{0.897}} \\
\hline
% \hline
&  DeepSVDD \cite{ruff2018deep}   & 0.622 & 0.597  & 0.560 & 0.509 & 0.563 & 0.581 & 0.583 & 0.549  &  0.513 \\
\rowcolor{gray!10} &+Multi-T  & 0.925 & 0.921 & 0.769  & 0.819 & 0.835 & 0.826 & 0.755 & 0.832  &  0.732  \\
\rowcolor{gray!10} & Improve.  & 48.7\% & 54.3\% & 37.3\% & 60.9\% & 48.3\% & 42.2\% & 29.5\% & 51.6\%  & 37.9\% \\
\hline
&  OCSVM  \cite{scholkopf2001estimating} & 0.927 & 0.921  & 0.826 & 0.850 & 0.879 & 0.850 & 0.826  & 0.871  & 0.831 \\
\rowcolor{gray!10} &+Multi-T   
& 0.957 & 0.965 & 0.859 & 0.916 & 0.916 & 0.899 & 0.846 & 0.924 & 0.863 \\
\rowcolor{gray!10} & Improve.  & 3.24\% & 4.78\% & 4.00\% & 7.76\% & 4.21\% & 5.76\% & 2.42\%
& 6.08\% & 3.85\%  \\
% \hline
% &  Shell-Re. \cite{lin2021shell} & 0.862 & 0.838 & 0.860 & 0.813 & 0.835 & 0.813 & 0.826 & 0.914  & 0.776 \\
% % \rowcolor{gray!10} & +Multi-T  & 0.959 & 0.976  & 0.875 & 0.931 & 0.926 & 0.917 & {{0.867}} & 0.947   &\\
% \rowcolor{gray!10} & +Multi-T  & \textcolor{blue}{0.968} & \textcolor{blue}{0.989} & \textcolor{blue}{0.895} & \textcolor{blue}{0.957} & \textcolor{blue}{0.938} & \textcolor{blue}{0.956} & \textcolor{orange}{0.867} & \textcolor{blue}{0.974} & \textcolor{blue}{0.897} \\
% \rowcolor{gray!10} & Improve. & 12.30\% & 18.02\% & 4.07\% & 17.71\% & 12.34\% & 17.59\% & 4.96\% & 6.56\% & 15.59\% \\
\bottomrule
% \hline
\end{tabular}}
\label{table 2}
\end{table}
}

\subsection{Outlier Scoring}
\label{sec 5.3}
\noindent {\textbf{Competing Methods.}}
We compare two themes of baseline outlier scoring (OS) methods: \noindent(\romannumeral1) statistical-based methods: IF \cite{liu2008isolation}, OCSVM \cite{scholkopf2001estimating}, DeepSVDD \cite{scholkopf2001estimating}, ECOD \cite{li2022ecod}, LUNAR \cite{lai2019robust}, RSRAE \cite{lai2019robust}, Shell-Re. \cite{lin2021shell}, LVAD \cite{lin2022locally}. \noindent (\romannumeral2) deep-learning-based models: GOAD \cite{bergman2020classification}, ICL \cite{shenkar2021anomaly}, REPEN \cite{pang2018learning}, NeuTraL \cite{qiu2021neural}, SLAD \cite{xu2023fascinating}. For a fair comparison, we apply Ergodic-set normalization \cite{lin2022locally} if it improves the performance of the baseline
algorithms, such as IF \cite{liu2008isolation}, OCSVM  \cite{scholkopf2001estimating}.
% Based on our above analysis, an effective threshold learner should advance the existing outlier detectors (scoring methods) and be effective on dataset split.
% The evaluation of the Multi-T module can be primarily divided into two aspects: results on thresholding tasks and the AUC improvement by leveraging Multi-T to previously established outlier scoring methods.

% Tab. \ref{tab 1}

\noindent
{\textbf{Evaluation Metric.}}
The performance of outlier scoring (ranking accuracy) is primarily assessed using the Area Under the Receiver Operating Characteristic curve (AUC). This metric provides a thorough assessment of ranking accuracy.

\noindent {\textbf{Main Results.}}
Tab. \ref{table 2} shows that Multi-T can be seamlessly integrated with two classic outlier scoring methods: DeepSVDD \cite{scholkopf2001estimating}, and OCSVM \cite{scholkopf2001estimating} while exhibiting significant improvements in both efficacy and stability across a diverse range of outlier ratios and various benchmarks. In most of our experiments, Multi-T itself achieves SOTA results. Notably, even in the case of non-aligned and low-resolution datasets like CIFAR-10 \cite{krizhevsky2009learning} and CIFAR-100 \cite{krizhevsky2009learning}, which are known to pose challenges \cite{perera2019ocgan}.
It surpasses the current SOTA AUC scores by margins of $7.76\%$ and $5.76\%$ for CIFAR-10 and CIFAR-100, respectively.
Additionally, we observe a logical and significant enhancement in performance with improved feature representation, e.g., for the MIT-Places \cite{zhou2017places} dataset, the AUC improves from $0.867$ using ResNet-50 \cite{he2016deep} to $0.974$ with CLIP \cite{radford2021learning}.

{
\renewcommand{\arraystretch}{1.2}
\begin{table}[htb!]
\centering
\footnotesize
\caption{Average AUC results, for more outlier scoring methods with(w/) or without(w/o) our Multi-T module, conducted on STL-10.} 
\setlength{\tabcolsep}{2.5pt}{
\begin{tabular}
{ll|cc|cc|cc|cc|ccc}
\toprule 
% \hline
& \multirow{2}{*}{{Feature}} &\multicolumn{2}{c}{IF \cite{liu2008isolation}} & \multicolumn{2}{c}{ECOD \cite{breunig2000lof}} & \multicolumn{2}{c}{ABOD \cite{kriegel2008angle}}  & \multicolumn{2}{c}{PCA \cite{shyu2003novel}} & \multicolumn{2}{c}{GMM \cite{aggarwal2016outlier}} \\
\cmidrule(lr){3-4} \cmidrule(lr){5-6} \cmidrule(lr){7-8} \cmidrule(lr){9-10} \cmidrule(lr){11-12}  
% \cline{3-12}
& & \makecell[c]{w/o} & \makecell[c]{w/} & \makecell[c]{w/o} & \makecell[c]{w/} & \makecell[c]{w/o} & \makecell[c]{w/} & \makecell[c]{w/o} & \makecell[c]{w/} & \makecell[c]{w/o} & \makecell[c]{w/}  \\
% \midrule
\hline
% \hline
& ResNet    & 0.836 & \cellcolor{gray!10} 
{0.899} & 0.907  & \cellcolor{gray!10}{0.919} & 0.665 & \cellcolor{gray!10}{0.883} & 0.865  &  \cellcolor{gray!10}{0.945} & 0.859  & \cellcolor{gray!10}{0.952}  \\ 
& CLIP     & 0.943 & \cellcolor{gray!10}{0.983} & 0.981  & \cellcolor{gray!10}{0.984} & 0.715   & \cellcolor{gray!10}{0.909} & 0.984  & \cellcolor{gray!10}{0.994}  & 0.892  & \cellcolor{gray!10}{0.962}  \\
& Avg.     & 0.890 & \cellcolor{gray!10}{0.941} & 0.944  & \cellcolor{gray!10}{0.951} & 0.690  & \cellcolor{gray!10}{0.896} & 0.925 &  \cellcolor{gray!10}{0.970} &  0.876 &  \cellcolor{gray!10}{0.957} \\ 
\hline
% \hline
& \multirow{2}{*}{{Feature}} &\multicolumn{2}{c}{GOAD \cite{bergman2020classification}} & \multicolumn{2}{c}{ICL \cite{shenkar2021anomaly}} & \multicolumn{2}{c}{REPEN \cite{pang2018learning}} & \multicolumn{2}{c}{NeuTraL \cite{qiu2021neural}} & \multicolumn{2}{c}{SLAD \cite{xu2023fascinating}} \\
\cmidrule(lr){3-4} \cmidrule(lr){5-6} \cmidrule(lr){7-8} \cmidrule(lr){9-10} \cmidrule(lr){11-12}   
% \cline{3-12}
& & \makecell[c]{w/o} & \makecell[c]{w/} & \makecell[c]{w/o} & \makecell[c]{w/} & \makecell[c]{w/o} & \makecell[c]{w/} & \makecell[c]{w/o} & \makecell[c]{w/} & \makecell[c]{w/o} & \makecell[c]{w/} \\
\hline  
% \midrule
& ResNet    & 0.952  & \cellcolor{gray!10}{0.962}  & 0.934  & \cellcolor{gray!10}{0.957}  & 0.877  &  \cellcolor{gray!10}{0.889}  & 0.854 & \cellcolor{gray!10}{0.950} & 0.941  & \cellcolor{gray!10}{0.962}   \\ 
& CLIP     & 0.961  & \cellcolor{gray!10}{0.989}  & 0.951  & \cellcolor{gray!10}{0.982}  & 0.879  & \cellcolor{gray!10}{0.930} & 0.851 & \cellcolor{gray!10}{0.971} &  0.945 & \cellcolor{gray!10}{0.985} \\ 
& Avg.     & 0.957  & \cellcolor{gray!10}{0.975}  & 0.943  & \cellcolor{gray!10}{0.970}  & 0.878  & \cellcolor{gray!10}{0.909} & 0.853 & \cellcolor{gray!10}{0.961} &  0.943 & \cellcolor{gray!10}{0.973} \\
\bottomrule
% \hline
\end{tabular}}
\label{table 3}
\end{table} 
}

\begin{figure*}[htb]
\centering
\includegraphics[width=0.99\textwidth]{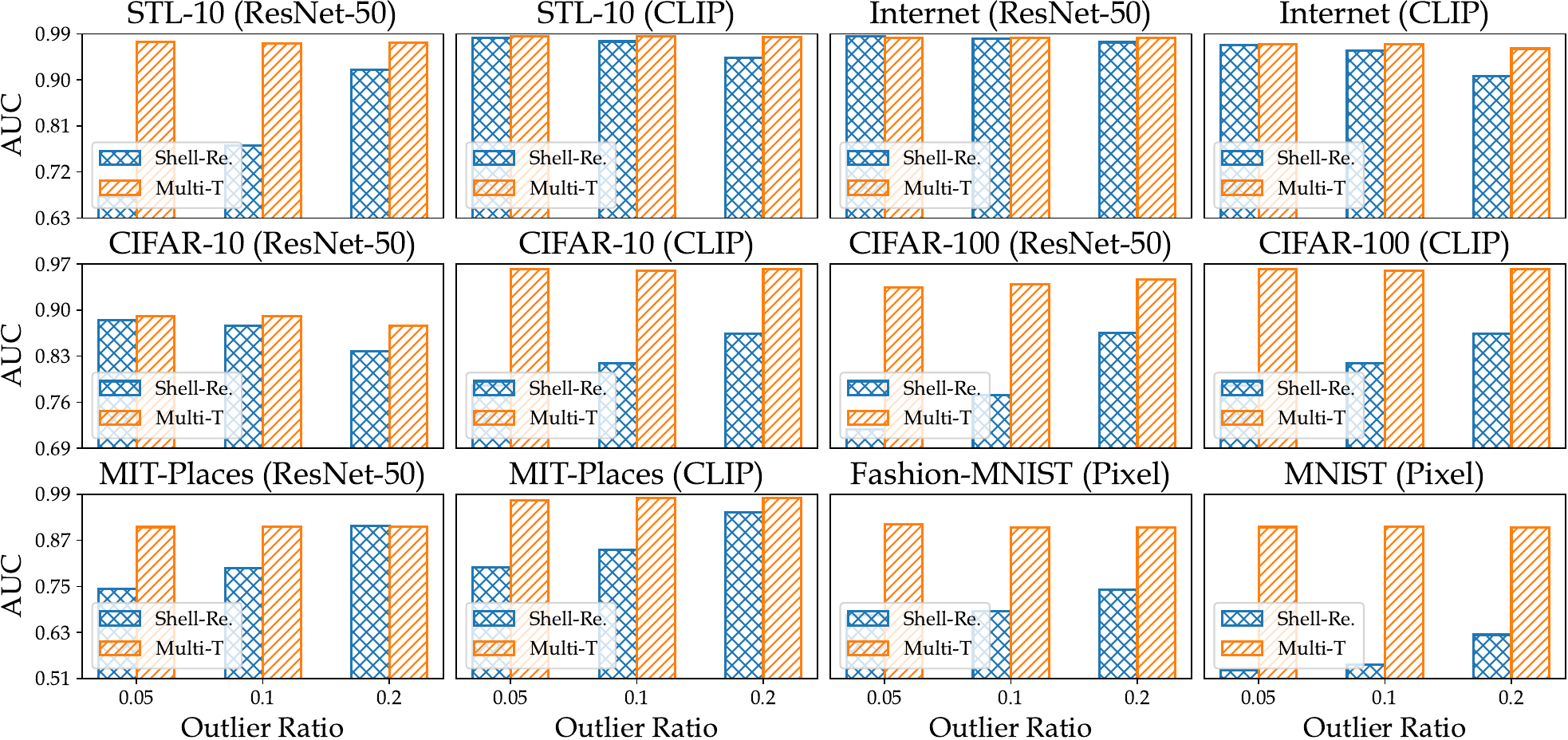}
\caption{AUC results of Multi-T compared with Shell-Re. \cite{lin2021shell} over various outlier ratios.
}
\label{figure 4}
 
\end{figure*}

In Tab. \ref{table 3}, we present the results of Multi-T integrated with more methods involving both statistical and deep models, which shows its broad applications.
Moreover, our solution excels not just in detection accuracy but also in running speed since Multi-T compresses the size of fitting data, as illustrated in Tab. \ref{table 4}, \ref{table 5}.
% More importantly, the high efficiency is still maintained, shown as Tab. \ref{table 3}.
In Fig. \ref{figure 4}, we compare with the most related work Shell-Re. \cite{lin2021shell}, which has a built-in Robust-Least-Square (RLS) thresholding procedure with MAD \cite{archana2015periodicity}. Apparently, our method is capable of estimating multiple thresholds that facilitate a clear demarcation of diverse scenarios (outlier ratio, dataset domain, and feature representation).

\begin{minipage}{\textwidth}
\begin{minipage}[t]{0.5\textwidth}
\makeatletter\def\@captype{table}
\renewcommand{\arraystretch}{1.1}
\centering
\footnotesize
\caption{Efficiency comparison for outlier scoring. Timing is measured with 10,000 samples, GPU: NVIDIA RTX 3080.}
\begin{tabular}{l|c|c}
    \toprule
    % \hline
    Method (OS)  & Device  & Time ($\downarrow$) \\
    \midrule
    % \hline
    LVAD \cite{lin2022locally}  & CPU  &   645.2s
    \\
    % ECOD \cite{li2022ecod}      &  7.0 
    % \\
    RSRAE \cite{lai2019robust} &  GPU  & 121.6s
    \\
    \rowcolor{gray!10} Multi-T  & CPU &  \textbf{1.2}s
    \\
    \hline
    % DeepSVDD   &  GPU & 26.6
    % \\
    % \rowcolor{gray!10} 
    % $+$Multi-T   &  $-$ &  25.1
    % \\
    OCSVM \cite{scholkopf2001estimating} & \multirow{2}{*}{CPU}   & 154.8s
    \\
% \rowcolor{gray!10} 
    \cellcolor{gray!10}$+$Multi-T  &  &  \cellcolor{gray!10}105.4s
    \\
    \hline
    
    GOAD \cite{bergman2020classification}  &  \multirow{2}{*}{GPU} &  268.1s
    \\
    % \rowcolor{gray!10} 
    \cellcolor{gray!10}$+$Multi-T   & &  \cellcolor{gray!10}211.7s
    \\
    \bottomrule
    % \hline
\end{tabular}
\label{table 4}
\end{minipage}
\begin{minipage}[t]{0.45\textwidth}
\makeatletter\def\@captype{table}
\centering
\footnotesize
\renewcommand{\arraystretch}{1.263}
\caption{Efficiency comparison for thresholding. All thresholding methods are conducted on CPU.}
\begin{tabular}{l|c}
    \toprule
    % \hline
    Method (TL)  & Time ($\downarrow$) \\
    \midrule
      CPD \cite{van2020evaluation}  &  3.61s  \\
      FWFM \cite{hashemi2019filtering} &  1.26s  \\
      DSN  \cite{amagata2021fast} &  3.79s  \\ 
      CLUST \cite{breunig2000lof}  &  7.36s  \\
      AUCP \cite{ren2018robust} &  1.26s   \\
    \hline
    \rowcolor{gray!10} 
    Multi-T   &  \textbf{0.97}s  \\
    \bottomrule
    % \hline
\end{tabular}
\label{table 5}
\end{minipage} 
\end{minipage}

\noindent {\textbf{Discussions.}}
We conclude the feasibility of adopting Multi-T to outlier scoring methods into three aspects: 
\noindent
(\romannumeral1) both normalization with \textit{outliers} and manifold learning with \textit{inliers} are necessary for most of outlier scoring methods; 
\noindent (\romannumeral2) Despite the initial outlier score distribution (Eq. \ref{equ 6}) is usually not perfectly separable, it provides a reliable baseline. 
\noindent (\romannumeral3) Multi-T is able to isolate the overlap region and predict both uncontaminated inliers.

\begin{minipage}{\textwidth}
\begin{minipage}[t]{0.5\textwidth}
\makeatletter\def\@captype{table}
\renewcommand{\arraystretch}{1.2}
\centering
\footnotesize
\caption{Ablation study for DeepSVDD and LVAD-S with Multi-T ($\gamma$: $0.2$).}
\begin{tabular}{lc|lc}
    \toprule
    % \hline
  \multirow{1}{*}{DeepSVDD }  & \multirow{1}{*}{AUC} & \multirow{1}{*}{LVAD-S} & \multirow{1}{*}{AUC} \\
    \midrule
    % \hline
     \rowcolor{gray!10}
    $+$Multi-T     &  \textcolor{blue}{0.950}     & $+$Multi-T  & \textcolor{blue}{0.971}  \\
    w/o In      &  0.643      & w/o In   & \textcolor{orange}{0.965} \\
    w/o Out      & \textcolor{orange}{0.939}    & w/o Out   & 0.951   \\
    \bottomrule
    % \hline
\end{tabular}
\label{table 6}
\end{minipage}
\begin{minipage}[t]{0.45\textwidth}
\makeatletter\def\@captype{table}
\renewcommand{\arraystretch}{1.15}
\centering
\footnotesize
\caption{Comparison with $k$-sigma.}
\begin{tabular}{c|l|c}
    \toprule
    % \hline
    \multirow{1}{*}{Metric} & \multirow{1}{*}{Method} & \multirow{1}{*}{Score} \\
    \midrule
    % \hline
    \multirow{2}*{\rotatebox{90}{$\text{F}_{0.1}$}} 
    % & 1-sigma     &    \\
    & 3-sigma    &  0.406  \\
    &\cellcolor{gray!10}Multi-T   &  \cellcolor{gray!10} \textbf{0.840}  \\
    \hline
    \multirow{2}*{\rotatebox{90}{$\text{F}_{10}$}}
    % & 1-sigma     &    \\
    & 1-sigma    &  0.806  \\
    &\cellcolor{gray!10}Multi-T   & \cellcolor{gray!10} \textbf{0.869}   \\
    \bottomrule
    % \hline
\end{tabular}
\label{table 7}
\end{minipage}
 
\end{minipage}

\subsection{Ablation Study}
% As the Multi-T module is plug-and-play and efficient, it can be integrated with any previously proposed outlier scoring systems.
% The experiments thus far have established the benefits of Multi-T.
% Given the efficacy of Multi-T,
% there remains a concern regarding 
The experiments thus far have established the effectiveness of Multi-T. However, there remains a concern regarding how sensitive the performance improvement is to Multi-T's two primary components: adaptive outliers (Out) and uncontaminated inliers (In), which refer to the normalization and manifold learning procedures, respectively.
We select two representative outlier scoring models: DeepSVDD \cite{ruff2018deep} and LVAD-S \cite{lin2022locally}, whereas
DeepSVDD follows a widely-used outlier detection mechanism, i.e., learning the normality (hyper-sphere), and identifying inliers is of great significance. Without predicted inliers, the AUC result decreases from $0.950$ to $0.643$ on STL-10, shown in Tab. \ref{table 6}. Additionally, since the normalization procedure can be considered as a distance de-noising procedure \cite{lin2023distance}, Multi-T will contribute to those methods with distance computation, e.g., the result of LVAD-S is improved from $0.951$ to $\mathbf{0.971}$ on STL-10 and $0.838$ to $\mathbf{0.882}$ on CIFAR-10, which verify our prior assumption that the identification of both inliers and outliers is of great significance, which enhances the reliability of of multiple thresholding perspective. 
Moreover, Tab. \ref{table 7} indicates that our thresholding process markedly surpasses the classical $k$-sigma rule \cite{pukelsheim1994three}. 
 
\subsection{Limitation} 
Our method is tied to the ranking accuracy (separability) of the initial outlier score function. Based on most related works, we consider UOD as an \textit{one-class} learning task (only one inlier/normal class in the target dataset). However we find in some specific cases, there might exist multi-normal classes, e.g., the digit 5 class of the MNIST dataset.
In that case, the efficacy of our method might be decreased. Additionally, this statistical-based method has limitations in very small-scale datasets since the "three-sigma" rule will be less effective.

\section{Conclusion}
This work introduces a novel perspective for UOD about advancing existing outlier detectors (scoring methods) via thresholding. To this end, we propose a multiple thresholding (Multi-T) module, to label the unlabelled target dataset. Comprehensive experiments
verify that the Multi-T can significantly improve both the efficacy and efficiency of previously proposed outlier scoring methods.
% We hope this work can inspire future
% research to propose better thresholding methods, rather than merely learning the outlier score function. Besides, Multi-T also isolates the overlap between inliers and outliers representing most inseparable samples, which is helpful for active learning tasks if supervision is allowed.

\section*{Acknowledegement}
% Wen-Yan Lin was supported by Lee Kong Chian Fellowship. 
We would like to thank the reviewers for their valuable comments and suggestions.

% ---- Bibliography ----
%
% BibTe\mathbf{x} users should specify bibliography style 'splncs04'.
% References will then be sorted and formatted in the correct style.
%
\bibliographystyle{splncs04}
% \bibliography{2763}

\end{document}